\pdfoutput=1

\documentclass[11pt]{article}

\usepackage[]{emnlp2021}

\usepackage{times}
\usepackage{latexsym}
\usepackage{makecell}
\usepackage{graphicx}
\usepackage{tikz}
\usepackage{multirow}
\usepackage{booktabs,tabularx,enumitem,ragged2e}
\usepackage{url}
\usepackage{amsmath}
\usepackage[normalem]{ulem}

\usepackage[T1]{fontenc}

\usepackage[utf8]{inputenc}

\usepackage{microtype}

\def\modelname{\textsc{VLC-StoryGAN}}

\title{Integrating Visuospatial, Linguistic and Commonsense Structure \\ into Story Visualization}

\author{Adyasha Maharana \quad \quad Mohit Bansal \\
    Department of Computer Science \\
  University of North Carolina at Chapel Hill \\
  {\tt \{adyasha, mbansal\}@cs.unc.edu} \\ \\}

\begin{document}
\maketitle
\begin{abstract}
While much research has been done in text-to-image synthesis, little work has been done to explore the usage of linguistic structure of the input text. Such information is even more important for story visualization since its inputs have an explicit narrative structure that needs to be translated into an image sequence (or visual story). Prior work in this domain has shown that there is ample room for improvement in the generated image sequence in terms of visual quality, consistency and relevance. In this paper, we first explore the use of constituency parse trees using a Transformer-based recurrent architecture for encoding structured input. Second, we augment the structured input with commonsense information and study the impact of this external knowledge on the generation of visual story. Third, we also incorporate visual structure via bounding boxes and dense captioning to provide feedback about the characters/objects in generated images within a dual learning setup. We show that off-the-shelf dense-captioning models trained on Visual Genome can improve the spatial structure of images from a different target domain without needing fine-tuning. We train the model end-to-end using intra-story contrastive loss (between words and image sub-regions) and show significant improvements in several metrics (and human evaluation) for multiple datasets. Finally, we provide an analysis of the linguistic and visuo-spatial information.\footnote{Code and data: \url{https://github.com/adymaharana/VLCStoryGan}.}
\end{abstract}

\begin{figure}[t]
\centering
\includegraphics[width=0.45\textwidth]{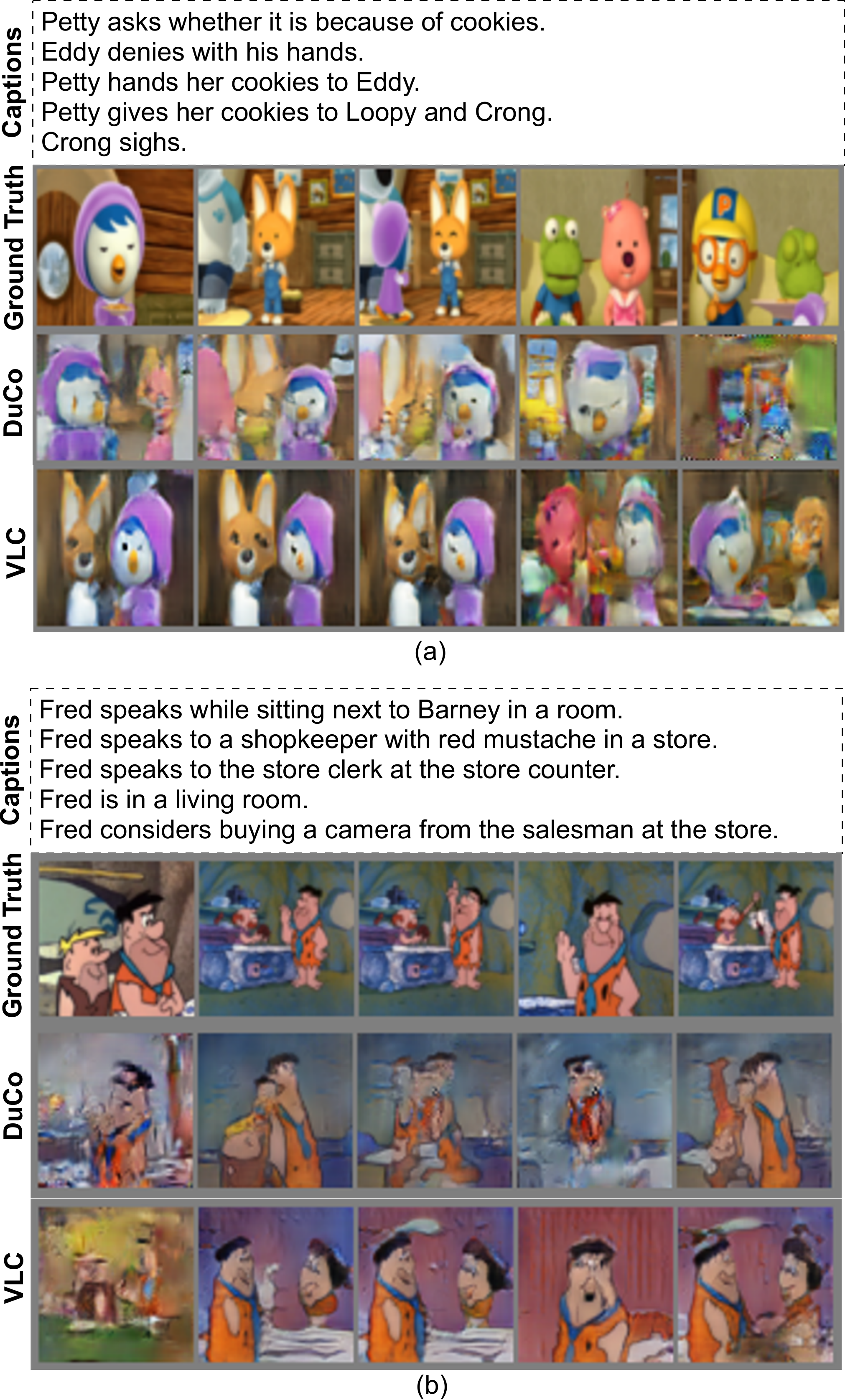}
\vspace{-8pt}
\caption{Example of generated images from our model \modelname{} and Duco-StoryGAN \cite{maharana2021ducostorygan} for the (a) PororoSV and (b) FlintstonesSV datasets.
\label{fig:pororo_example}}
\vspace{-12pt}
\end{figure}

\section{Introduction}
\label{sec:intro}
Story Visualization is an emerging area of research with several potentially interesting applications such as visualization of educational materials, assisting artists with web-comic creation etc. Each story consists of a sequence of images along with a sequence of captions describing the content of the images. The goal of the task is to reproduce the images given the captions. It is more challenging than conventional text-to-image generation \cite{reed2016generative} because the generative model needs to identify the narrative structure expressed in the sequence of captions and translate it into a story of images. Some critical features of a good story include consistent character and background appearances, relevance to individual captions as well as overall story, and coherent narrative. While recent text-to-image models \cite{ramesh2021zero, cho2020x, li2019controllable} are successfully generating high-quality images, they are not directly designed for narrative understanding over sequential text. Hence, story visualization necessitates independent research towards developing generative models for the task. In this paper, we explore the use of visuo-linguistic structured inputs and outputs for improving story visualization. Towards this end, we propose (V)isuo-spatial, (L)inguistic \& (C)ommonsense i.e. \modelname{} which (1) uses constituency parse trees and commonsense knowledge as input using structure-aware encoders, (2) leverages a pretrained dense captioning model for additional position and semantic information, and (3) is trained using intra-story contrastive loss for maximizing global semantic alignment between input captions and generated visual stories.

Grammatical structures like constituency parse trees are potentially rich sources of information for visualizing relations between objects (or characters), their actions, and their attribute (property) modifiers. \citet{wang2019tree, nguyen2020tree, xiao2017weakly, cirik2018using} demonstrate that inducing such tree-structures within the encoder guides words to compose the meaning of longer phrases hierarchically and improves various tasks like masked language modeling, translation, visual grounding of language etc., suggesting potential gains in other tasks. Most text-to-image synthesis as well as story visualization models \cite{li2019storygan, maharana2021ducostorygan} perform flat processing over free-text captions using LSTM or Transformer-based encoders. Hence, in order to leverage the grammatical information packed in constituency parse trees, we propose a novel Memory-Augmented Recurrent Tree-Transformer (MARTT) to encode captions and promote forward flow of hierarchical information across the sequence of captions for each story. Further, we find that input captions in story visualization lack details about the visual elements in the image. Hence, we augment the captions with external knowledge. For instance, when one caption mentions \textit{snow} while the other mentions \textit{icy roads}, we provide the knowledge that both are related to cold weather, encouraging the model to learn similar representations for either of the phrases. 

Dual learning has served as an effective method for promoting desirable characteristics in target output for both text-to-image generation \cite{qiao2019mirrorgan} and story visualization \cite{maharana2021ducostorygan}. \citet{song2020CPCSV} use image segmentation to preserve character shapes while \citet{maharana2021ducostorygan} use video captioning for global alignment between the input caption and the generated sequence of frames. Each of these auxiliary tasks generate uni-modal outputs, dealing either with image or text. In a bid to combine the benefits of learning signals from both visuo-spatial and language modalities, we propose the use of dense captioning as the dual task, which has proven useful as a source of complementary information for many vision-language tasks \cite{wu2019generating, kim2020dense, li2019visual}. Dense captioning models provide regional bounding boxes for objects in the input image and also describe the region. By using these outputs for dual learning feedback for story visualization, the generative model receives a signal rich in spatial as well as semantic information. The spatial signal is especially important for our task since the input captions do not contain any specifications about the shape, size or position of characters within the story. Further, we find that off-the-shelf dense captioning models, which are trained on realistic images from Visual Genome, transfer well to a markedly different domain like cartoon and can be used to provide visuo-spatial feedback without finetuning on target domain.

Finally, we want the model to recognize the subtle differences between frames in a story and generate relevant images that fit into a coherent narrative. Hence, we employ contrastive loss between image-regions and words in the captions at each timestep to improve semantic alignment between the caption and image. Adjacent frames in a story often contain subtle differences, as can be seen in an example in Fig.~\ref{fig:pororo_example}. We modify the region-word contrastive loss proposed in \citet{zhang2021cross} for story visualization by sampling negative images from adjacent frames, forcing the model to recognize the difference between frames. 

Overall, our contributions are: (1) We propose \modelname{} to use linguistic information, augmented with commonsense knowledge, for conditional image synthesis. (2) use dense-captioning to provide complementary positional and semantic information during training and show that off-the-shelf models trained on Visual Genome can be effective without fine-tuning on the target domain. (3) propose intra-story contrastive loss between image regions and words to improve semantic alignment between captions and visual stories. (4) achieve strong improvements in several metrics and human evaluation for multiple datasets (PororoSV and FlintstonesSV), compared to previous state-of-art, and show the usefulness of structured inputs and outputs to provide insights for future work.

\section{Related Work}
\paragraph{Story Visualization.} The task of story visualization and the model StoryGAN was introduced by \citet{li2019storygan}. \citet{zeng2019pororogan} and \citet{LI2020102956} used textual alignment modules and Weighted Activation Degrees respectively, to improve performance of StoryGAN. \citet{song2020CPCSV} add a figure-ground generator and discriminator to preserve the shape of characters. \citet{maharana2021ducostorygan} demonstrate the effectiveness of video captioning as a dual task for story visualization and propose additional evaluation metrics. Notable recent models in the related field of text-to-image generation are large \cite{brock2018large}, trained on gigantic datasets \cite{ramesh2021zero} and are based on Transformer architectures \citet{jiang2021transgan}. Mask-to-image generation modules \cite{koh2021text} have proven effective for smaller datasets containing detailed captions and additional information for aligning image sub-regions to words within captions \cite{pont2020connecting}. This is in sharp contrast to the datasets available for story visualization, which have been repurposed from video QA datasets and hence, contain short descriptions. Our work is based on exploring structured inputs and outputs for conditional image synthesis which has been largely unexplored in text-to-image synthesis and story visualization.

\paragraph{Story Understanding \& Commonsense.} \citet{iyyer2016feuding} introduced Relationship Modelling Networks to extract evolving relationship trajectories between two characters in a novel. \citet{chaturvedi2017story} use latent variables to weigh pre-defined semantic aspects like topical consistency to improve encoding for story completion. \citet{guan2019story, chen2019incorporating} augment story encodings with structured commonsense knowledge to improve story ending generation. We focus on the use of structured commonsense as well as grammatical trees to improve story encoding for the end goal of visualization.

\paragraph{Tree Encoder.}
Tree structures have traditionally been encoded using Tree LSTMs \cite{tai2015improved, miwa2016end, yang2017dense, yang2017towards}. In recent work, \citet{wang2019tree} enforce a hierarchical prior in the self-attention layer of Transformer \cite{vaswani2017attention} and \citet{harer2019tree} use a parent-sibling tree convolution block to perform structure-aware encoding. \citet{nguyen2020tree} use sub-tree masking and hierarchical accumulation to improve machine translation. We propose a simpler Tree-Transformer architecture, augmented with memory units \cite{lei2020mart} for recurrence.

\paragraph{Contrastive Loss.} \citet{xu2018attngan} first proposed the contrastive loss in text-to-image synthesis through the Deep Attentional Multimodal Similarity Model (DAMSM). ContraGAN \cite{kang2020contragan} performs minimization of contrastive loss between multiple image embeddings in the same batch, in addition to class embeddings \cite{miyato2018cgans}. \citet{zhang2021cross} combine inter-modality and intra-modality contrastive losses and observe complementary improvements. We adapt inter-modal loss for story visualization by sampling negatives from adjacent frames.

\paragraph{Dense Captioning.} Dense captioning jointly localizes semantic regions and describes these regions with short phrases in natural language \cite{johnson2016densecap}. \citet{wu2019generating} and \citet{kim2020dense} use dense captions for visual and video question answering respectively. We use a pretrained dense captioning model to first annotate our target dataset and then use it within a dual learning framework to improve image synthesis for story visualization.

\begin{figure*}[t]
\centering
\includegraphics[width=0.99\textwidth]{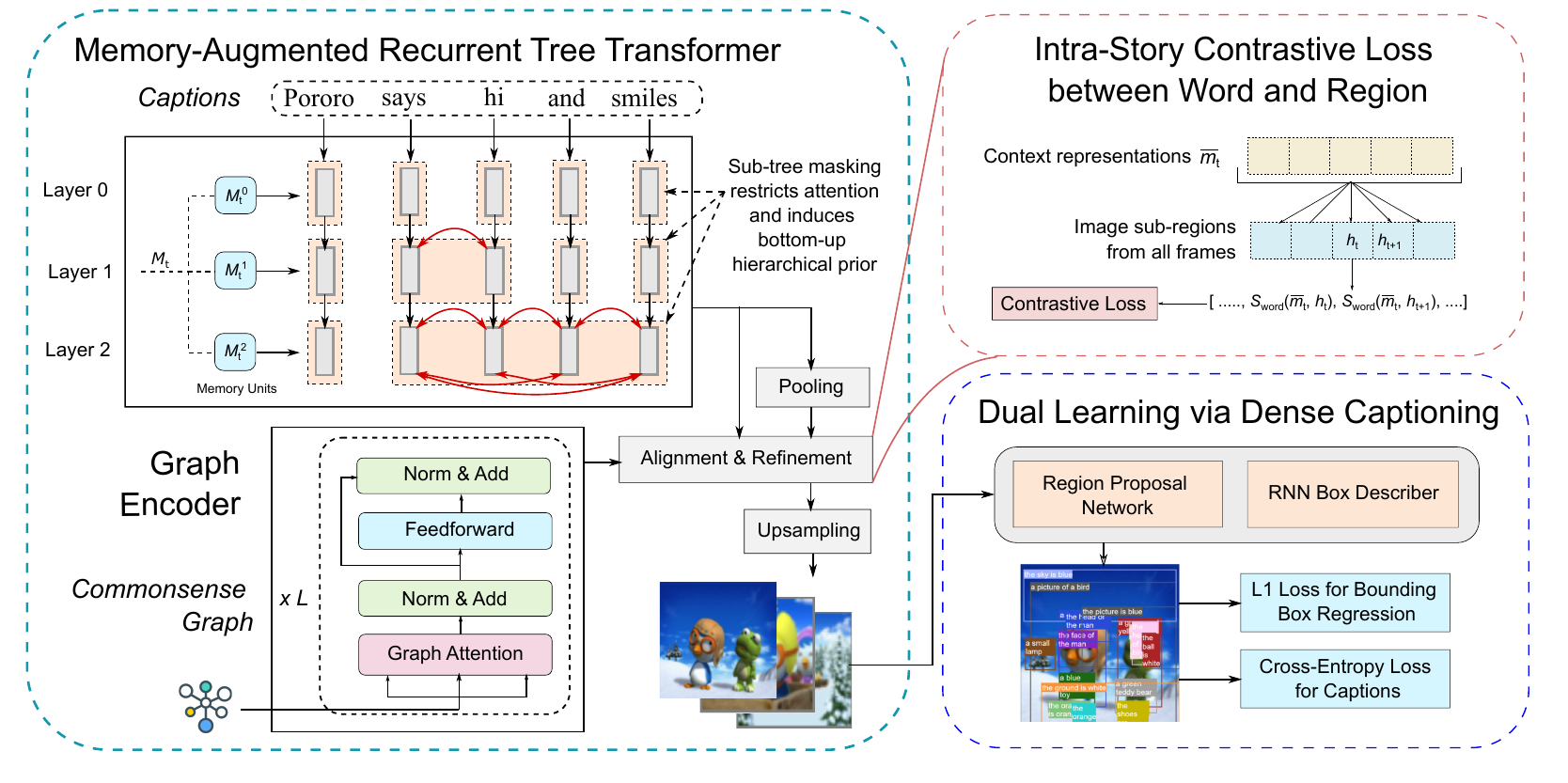}
\vspace{-9pt}
\caption{Illustration of \modelname{} architecture. The story encoder is composed of MARTT for encoding sequence of constituency parse trees, and Graph Transformer for encoding commonsense knowledge graphs. The intra-story contrastive loss optimizes semantic alignment while dense captioning loss provides visuo-spatial and semantic feedback about object/characters in generated images.\vspace{-3pt}
\label{fig:model_architecture}}
\vspace{-1pt}
\end{figure*}

\begin{figure}[t]
\centering
\includegraphics[width=0.49\textwidth]{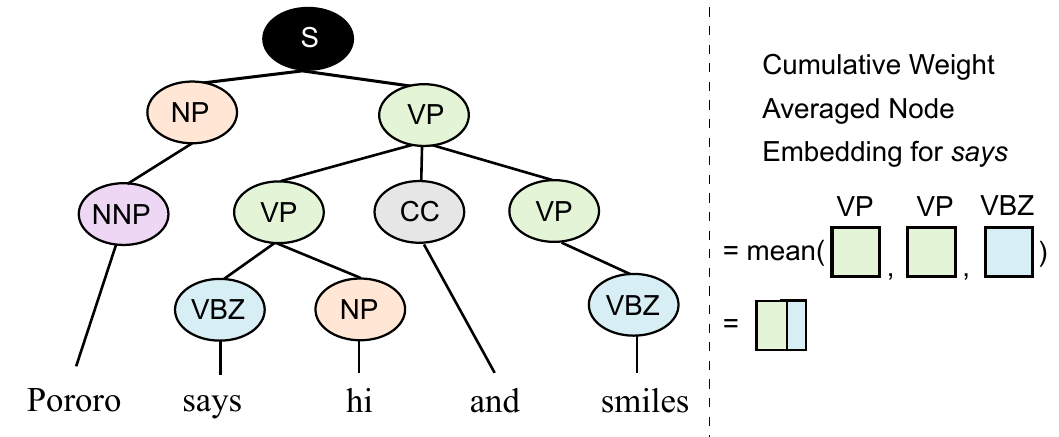}
\vspace{-7pt}
\caption{Upward cumulative avg. of node embeddings. \label{fig:node}}
\vspace{-20pt}
\end{figure}

\section{Methods}
\subsection{Background}
Given a sequence of sentences $S=[s_1, s_2, ..., s_T]$, story visualization is the task of generating a corresponding sequence of images $\hat{X} = [\hat{x}_1, \hat{x}_2, ..., \hat{x}_T]$. The sentences form a coherent story with recurring plot and characters.  The generative model for this task has two main modules: story encoder and image generator. The story encoder $E(.)$ consists of a recurrent encoder which takes word embeddings $\{w_{ik}\}$ for sentence $s_k$ at each timestep $k$ and generates contextualized embeddings $\{c_{ik}\}$. $E(.)$ also learns a stochastic mapping from $S$ to a representation $h_0$ which encodes the whole story and is used to initialize hidden states of the recurrent encoder \cite{li2019storygan, maharana2021ducostorygan}. The image generator $I(.)$ takes $\{c_{ik}\}$ and pools them into representations $\{o_k\}$ which are then transformed into images $\{\hat{x}_k\}$. We train the model within a GAN framework \cite{goodfellow2014generative}. The generated images are passed to image and story discriminators, which evaluate the images in different ways and send back a learning signal. 

In \modelname{}, we use constituency trees as input to a structure-aware encoder. Further, we impose losses based on visuo-linguistic structures and contrastive loss on the model during training. We outline each of these modules in detail.

\subsection{Memory-Augmented Recurrent Tree Transformer (MARTT)}

Given a sentence $s$ of length $n$, let $G(s)$ be the constituency parse tree of $s$ produced by a parser. $T(s)$ denotes the ordered sequence of $n$ terminal nodes (or leaves) of the tree and $N(s)$ denotes the set of nonterminal nodes (or simply nodes), each of which has a phrase label (e.g., NP, VP) and spans over a sequence of terminal nodes. Each leaf embedding is the concatenation of word embedding and a corresponding node embedding. To compute node embeddings, we perform upward cumulative average \cite{nguyen2020tree} over the nodes (phrase labels) that the respective non-terminal leaf token is a child of. For instance, as seen in Fig.~\ref{fig:node}, the node embedding for the word \textit{Pororo} is the average of embeddings for \textsc{NNP} and \textsc{NP}. The node representations are learnt during training and provide information about the phrase label classes for each token. The encoder receives the sequence of leaf embeddings, in the same order as in the sentence, as input. Within the encoder, the hierarchical structure of a parse tree is promoted by introducing sub-tree masking for encoder self-attention \cite{wang2019tree}. For each word query, self-attention has access only to other members of the sub-tree at that layer. In Fig.~\ref{fig:model_architecture}, each token only attends to itself in the first layer of Tree-Transformer. In the next layer, \textit{says} and \textit{hi} can attend to each other as they belong to the sub-tree rooted at \textsc{VP}. Consequently, all tokens within \textit{says hi and smiles} can attend to each other in the third layer. This bottom-up approach, paired with node embeddings, induces the model to build a hierarchical understanding of the sentence through compositionality.

Tree Transformer was originally designed to encode a single tree input whereas in our task, we need to encode a sequence of trees for the sequence of images we plan to generate. Hence, we tie a series of Tree Transformers together by introducing memory cells and memory updater modules in each layer of self-attention. At time step $t$, the input query matrix within the self-attention layer attends over $[M_{t-1}^{l};\bar{H}_{t}^{l}]$ where $M\in R^{T_{m}\times d}$ and $\bar{H}\in R^{T_{c}\times d}$ ($T_m$ denotes memory state length and $T_c$ denotes length of caption). The memory state $M_{t-1}^{l}$ is updated to $M_{t}^{l}$ following the steps outlined for memory updater in \citet{lei2020mart}.

\subsection{Commonsense Knowledge}
The input captions in most narrative datasets generally omit several relevant details about the plot or the background, which can be considered as commonsense. For example, in a scene where two characters are present outside on a sunny day, the caption does not explicitly mention the presence of a sky in the background or the brightness of the sun. Hence, in order to introduce this external knowledge and enrich the input captions, we extract commonsense concepts relevant to each frame. To do so, we follow \citet{bauer2018commonsense} and use a simple entity-based method to extract relevant paths from ConceptNet (see Sec.~\ref{sec:experiments} for details). 

The commonsense knowledge paths are merged into a sub-graph which is then encoded using Graph Transformer. We use the Transformer-based graph encoder from Graph Writer \cite{koncel2019text} for structure-preserving encoding of graphs. First, the input graphs $g_{k}$ are converted into unlabeled connected bipartite graphs $G_{k} = (v_{k}, E_{k})$, where $v_{k}$ is the list of entities and relations, and $e_{k}$ is the adjacency matrix describing the directed edges \cite{beck2018graph}. Next, $v_{k}$ is projected to a dense, continuous embedding space $V_{k}$ and is sent as input to the graph encoder. The encoder is composed of $L$ stacked Transformer blocks; each Transformer block consists of a $N$-headed self-attention layer followed by normalization and a two-layer feed-forward network. The resulting encodings are referred to as graph contextualized vertex encodings. The entity encodings $e_k$ are then appended to the output $c_k$ from MARTT and used in the alignment module (see Fig.~\ref{fig:model_architecture}).

\subsection{Image Generation}
\label{sec:decoder}
The image generator follows the two-stage approach in prior text-to-image generation works \cite{qiao2019mirrorgan,xu2018attngan,han2017stackgan, maharana2021ducostorygan}. The alignment module performs attention-based semantic alignment \cite{xu2018attngan} between image regions $h_k$ and words $\bar{m}_{k}=[f_{entity}(e_{k});f_{caption}(c_{k})]$ in the current timestep. $f_{entity}$ and $f_{caption}$ are dense layers for projecting commonsense and caption encodings respectively, into the same space as image embeddings. $\beta_{jik}$ indicates the weight assigned by the model to the $i^{th}$ word when generating the $j^{th}$ sub-region of the image. For the $j^{th}$ image sub-region, the word-context vector is calculated as:
\begin{gather*}
    a_{jk} = \sum_{i=0}^{L}\beta_{ji}\bar{m}_{ik};\>\> \beta_{jik}=\frac{\text{exp}(h_{jk}^{T}\bar{m}_{ik})}{\sum_{i=0}^{L}\text{exp}(h_{jk}^{T}\bar{m}_{ik})}
\end{gather*}

The generated images are sent to image and story discriminators and the corresponding classification loss is used for training. We use the discriminator models proposed in \citet{li2019storygan}. Given the sentence $s_k$ and the context information vector from the story encoder $h_0$, the image discriminator attempts to distinguish between the generated and ground truth image $x_k$, resulting in the loss $\mathcal{L}_{img}$. Similarly, the story discriminator classifies between the ground truth story and the generated sequence of images $\hat{X}$ to produce the loss $\mathcal{L}_{story}$. Additionally, the image discriminator is also used to classify the characters in the frame, when labels are available.

\subsection{Dual Learning with Dense Captioning} As we discussed in Sec.~\ref{sec:intro}, dual learning can provide important visual or semantic signals for improving story visualization, depending on which auxiliary task is chosen for the feedback model. We propose the use of dense captioning for providing visuo-spatial as well as semantic learning signals during training and use the model in \citet{yang2017dense} as the feedback model.\footnote{We use the implementation at \url{https://github.com/soloist97/densecap-pytorch}} The dense captioning model is not fine-tuned on images from the story visualization dataset since it lacks dense caption annotations and it is prohibitively time-consuming and expensive to gather such annotations for the task. Hence, we explore the use of Visual Genome-based predictions \cite{krishna2017visual} as "proxy" annotations for our dataset (see Fig.~\ref{fig:model_architecture}). Using these noisy predictions as ground-truth, we train the generative model to optimize for bounding box loss (L1 regression; $\mathcal{L}_{bbox}$) as well as captioning loss (cross-entropy; $\mathcal{L}_{caption}$).

\noindent\textbf{Position Invariance via Bounding Box Loss: } The input captions in our dataset do not specify positions for the characters. Unless there is explicit positional input, it is unreasonable to expect the model to get the ground truth positions correct in generated images. Hence, in order to enforce positional invariance, we augment our dataset with mirror versions of the frames.

\subsection{Contrastive Loss}
As discussed in Sec.~\ref{sec:decoder}, the alignment and refinement module computes a pairwise cosine similarity matrix between all pairs of image-regions and word tokens, followed by the soft attention $\beta_{i,j}$ for image region $j$ to word $i$. The aligned word-context vector $a_{j}$ for the $j^{th}$ sub-region is the weighted sum of all word representations. Following \citet{zhang2021cross}, the score function between all sub-regions $h_k$ for image $x_k$ and all words $\bar{m}_k$ corresponding to caption $s_k$ is defined as $S_{word}(x_k, s_k) = log(\sum_{j=1}^N exp(cos(h_{jk}, a_{jk})))$, where $N$ is the total number of sub-regions. Finally the contrastive loss between the words and regions in image $x_k$ and its aligned sentence $s_k$ with respect to the story is defined as:
\begin{equation*}
    \mathcal{L}_{word} = -log \frac{exp(S_{word}(x_k, s_k))}{\sum_{m=1}^T exp(S_{word}(x_m, s_k))}
\end{equation*}
where $T$ is the total number of frames in a story.

\paragraph{Conditioning Mechanism.} The story encoder $E(.)$ encodes the entire story $S$ into a single representation, $h_0$, which functions as the initial memory state of the MARTT model, similar to \citet{maharana2021ducostorygan}. The input $S$ is the concatenation of sentence embeddings $s_k\in\mathcal{R}^{1\times d_s}$ from all timesteps. The conditional augmentation technique \cite{han2017stackgan} is used to convert $S$ into a conditioning vector by using it to construct and sample a conditional Gaussian distribution i.e., $h_0 = \mu(S) + \sigma^2(S)^{1/2} \odot \epsilon_{S}$, where $\epsilon_{S} \sim \mathcal{N}(0, 1)$ and $\odot$ represents element-wise multiplication. This introduces a KL-Divergence loss between the learned distribution and Gaussian distribution i.e., $\mathcal{L}_{KL} = KL(\mathcal{N}(\mu(S), \text{diag}(\sigma^2(S))) || \mathcal{N}(0, I))$.

\paragraph{Objective.} The final objective function of the generative model is $\min_{\theta_{G}} \max_{\theta_{I},\theta_{S}} [\mathcal{L}_{KL} + \mathcal{L}_{img} + \mathcal{L}_{story} + \lambda_{bbox}\mathcal{L}_{bbox} + \lambda_{caption}\mathcal{L}_{caption} + \mathcal{L}_{word}]$ where $\theta_{G}$, $\theta_{I}$ and $\theta_{S}$ denote the parameters of the entire generator, and image and story discriminator respectively. $\lambda$ values are weight factors for the respective losses.

\section{Experimental Settings}
\label{sec:experiments}
\paragraph{Evaluation.} We adopt the metrics proposed in \citet{li2019storygan} and \citet{maharana2021ducostorygan}:
\begin{itemize}[leftmargin=*]
\itemsep0em 
    \item \textbf{Character Classification}: Frame accuracy (exact match) and classification F1-score using finetuned Inception-v3 to measure visual quality of recurring characters in predicted images. \cite{szegedy2016rethinking}.
    \item \textbf{Video Captioning Accuracy}: BLEU2/3 scores of captions generated for predicted images using pretrained MART \cite{lei2020mart}.
    \item \textbf{R-Precision}: R-Precision for global semantic alignment between predicted images and groud truth captions using the Hierarchical-DAMSM \cite{maharana2021ducostorygan}.
    \item \textbf{Frechet Inception Distance (FID)}: The distance between distributions of real images and generated images using pretrained Inception-v3.

\end{itemize}

Since story visualization datasets are adapted from a video captioning dataset, sometimes a single frame does not represent the caption perfectly. However, during training, we sample a frame from the video every time, thus providing coverage for the entire video and association between all characters in the story and their representation in the frame. With this process, the model is able to observe all characters from the caption in the target frames during training time. During inference, our target is a static story, and not a video. Hence, we evaluate the predictions under the assumption that all characters should appear in the frame.

\renewcommand{\arraystretch}{1.5}%
\begin{table*}[t]
\small
\centering
\begin{tabular}{|l|c|c|c|c|c|}
\hline
\textbf{Model}  & \textbf{Char. F1} & \textbf{Frame Acc.} & \textbf{FID$\downarrow$} & \textbf{BLEU2/3} & \textbf{R-Precision} \\ 
\hline
StoryGAN \cite{li2019storygan}  & 18.59 & 9.34 & 49.27 & 3.24 / 1.22 &  1.51 $\pm$ 0.15 \\
CP-CSV \cite{song2020CPCSV}  & 21.78 & 10.03 & 40.56 & 3.25 / 1.22 & 1.76 $\pm$ 0.04 \\
\textsc{Duco-StoryGAN} \cite{maharana2021ducostorygan}  & 38.01 & 13.97 & 34.53 & 3.68 / 1.34 &  \textbf{3.56 $\pm$ 0.04} \\
\modelname{} (Ours)  & \textbf{43.02} & \textbf{17.36} & \textbf{18.09} & \textbf{3.80 / 1.44} & 3.28 $\pm$ 0.00  \\
  
\hline
\end{tabular}
\caption{\label{tab:pororo_test} Results on test split of PororoSV Dataset. Lower FID is better; higher is better for rest of the metrics.}
\vspace{-7pt}
\end{table*}

\paragraph{Dataset.} We use the PororoSV dataset proposed in \citet{li2019storygan}, and the splits proposed in \citet{maharana2021ducostorygan} to evaluate our approach. Each sample in PororoSV has 5 frames and 5 corresponding captions that form a narrative. There are 9 recurring characters throughout the dataset. Each character is featured in at least 10\% of the frames, making it crucial for the model to be capable of generating each of them. There are 10191/2334/2208 samples in training, validation and test splits respectively. The constituency parses are extracted and pre-processed using spaCy~\cite{kitaev2018constituency} and NLTK \cite{bird2009natural}.\footnote{\url{https://spacy.io/universe/project/self-attentive-parser}} For commonsense knowledge, we first extract nouns and verb words from all of the captions in a story, and find ConceptNet triples \cite{speer2017conceptnet} containing at least one of those words in the subject and object phrases. Next, we use pretrained GloVe embeddings \cite{pennington2014glove} to find a broader pool of words which are related to the words and find additional relevant triples. These triples are combined into knowledge graph inputs for each frame. We use the top ten bounding box and caption predictions from a dense captioning model pretrained on Visual Genome \cite{krishna2017visual} for dual learning.

\paragraph{Experiments.}
Our model is developed using PyTorch. All models are trained on the proposed training split and evaluated on validation and test sets. We select the best checkpoints and tune hyperparameters by using the character classification F-Score on the validation set.

\begin{table}[t]
\small
\centering
\begin{tabular}{|c|c|c|c|}
\hline
Attribute & Win\% & Lose\% & Tie\% \\ 
\hline
Visual Quality & 62\% & 28\% & 10\% \\
Consistency & 38\% & 30\% & 32\% \\
Relevance & 22\% & 18\% & 60\% \\
\hline
\end{tabular}
\caption{\label{tab:human_eval} Results from human evaluation. Win\% = \% times stories from \modelname{} was preferred over DuCo-StoryGAN, Lose\% for vice-versa. Tie\% represents remaining samples.}
\end{table}

\section{Results}

\subsection{Main Quantitative Results}
The results on the PororoSV test set can be seen in Table \ref{tab:pororo_test}. We compare our model \modelname{} to three baselines: StoryGAN \cite{li2019storygan}, CP-CSV \cite{song2020CPCSV} and \textsc{DuCo-StoryGAN} \cite{maharana2021ducostorygan} for PororoSV. The final rows contain results with \modelname{}, which outperforms previous models across most metrics for PororoSV. We see drastic improvements in FID score and sizable improvements in charcater classification as well as frame accuracy scores. This demonstrates the superior visual quality of stories visualized via our proposed method. There is a small improvement in BLEU score and a slight drop in R-Precision.

The captions in PororoSV correspond more accurately to a video segment than a single image sampled from the segment (see example in Fig.~\ref{fig:pororo_example}). Hence, even though the metrics BLEU and R-Precision have been shown to be correlated with human judgement in text-to-image synthesis \cite{hong2018inferring}, the PororoSV dataset fails to be an appropriate testing bed for extending those metrics to story visualization. Since they are adapted from video datasets, there is poor correlation between a single frame and the caption that originally spanned an entire video clip. This leads to unstable results and smaller improvement margins for both metrics. Instead, the dataset presents a data-scarce scenario where the captions do not provide sufficient details for accurate generation of visual stories. This leaves ample scope for augmenting the input with external visual information such as scene graphs and dense captions, or structured knowledge such as commonsense graphs, as we have shown with our proposed model. The structured information in \modelname{} leads to better generation of multiple characters, as compared to Duco-StoryGAN (Fig.~\ref{fig:pororo_example}).

\renewcommand{\arraystretch}{1.5}%
\begin{table*}[t]
\small
\centering
\begin{tabular}{|l|c|c|c|c|c|}
\hline
\textbf{Model}  & \textbf{Char. F1} & \textbf{Frame Acc.}  & \textbf{FID$\downarrow$} & \textbf{BLEU2/3} & \textbf{R-Precision}  \\ 
\hline
\modelname{} & 50.07 & 25.33 & 18.08 & 4.57 / 2.14 & 6.06 $\pm$ 0.00 \\
\hline 
- MARTT & 48.96 & 22.84 & 24.56 & 4.12 / 1.59 & 5.86 $\pm$ 0.01 \\
- Commonsense Embeddings  & 50.02 & 25.17 & 18.41 & 4.57 / 2.10 & 6.08 $\pm$ 0.02 \\
- Dense Captioning & 49.87 & 24.68  & 20.02 & 4.32 / 1.84 & 6.07 $\pm$ 0.00 \\
- Intra-Story Contrastive Loss  & 48.65 & 24.98 & 21.67 & 4.43 / 1.92 & 5.99 $\pm$ 0.01  \\
\hline
\end{tabular}
\caption{\label{tab:pororo_ablation} Ablation results for our model on validation split of PororoSV dataset. Lower FID indicates better performance, higher is better for rest of the metrics. Dense captioning includes both bounding box and captioning.}
\vspace{-5pt}
\end{table*}

\subsection{Human Evaluation}
We conduct human evaluation on the generated images from \modelname{} and DuCo-StoryGAN, using the three evaluation criteria listed in \citet{li2019storygan}: visual quality, consistence, and relevance (see Appendix for details). Predictions from our model for PororoSV are preferred 62\% of the times for better visual quality (see Win\% columns). Our model also produces more consistent and relevant images, but the higher \% of ties between the two models for these attributes indicate that much work remains to be done to improve global alignment between captions and images.

We also examine 50 random samples from the PororoSV dataset, and evaluate whether the bounding boxes predicted by the pretrained dense captioning model used in our approach are relevant to the task i.e. whether more than 50\% of the predicted bounding boxes for each sample capture a meaningful part of the frame. We observe a high accuracy for PororoSV i.e. 68\%.

\subsection{Ablations}
Table \ref{tab:pororo_ablation} contains minus-one ablations for \modelname{} on the PororoSV validation set. The first row shows results from the complete model \modelname{}. We then iteratively remove each of our contributions and observe the change in metrics. We obtain the largest drops in FID, character classification and frame accuracy by replacing MARTT with the structure-agnostic MART (second row). This suggests that the constituency tree, as well as the MARTT architecture, aids in comprehension of captions. We see similar but smaller drops with the exclusion of dense captioning from \modelname{}, since it provides important positional and semantic information about visual elements (third row). The minor margins for commonsense knowledge (fourth row) suggest that while it is a promising source of additional data, more work is needed for its proper integration with input captions. Finally, the results in the last row show that the intra-story contrastive loss is effective for global semantic alignment.\footnote{On the validation set of PororoSV, \modelname{} outperforms previous models across all metrics except FID score, where StoryGAN has the best FID score. This may be attributed to the 58\% frame overlap between training and validation sets of PororoSV, since this trend does not transfer to the test set (where \modelname{} is best on FID) which has zero overlap with the training split. The FID metric has also been shown to be biased for finite sample estimates \cite{chong2020effectively}.}

We also ran an experiment for isolating the effect of memory augmentation in our model, by training a non-recurrent (no memory) Transformer with Tree representations for single image generation instead of story generation, and evaluated using the story visualization metrics. We observed significant drops across all metrics.

\section{Analysis and Discussion}

In this section, we take a closer look at the various data sources for \modelname{}.

\begin{figure}[t]
\centering
\includegraphics[width=0.47\textwidth]{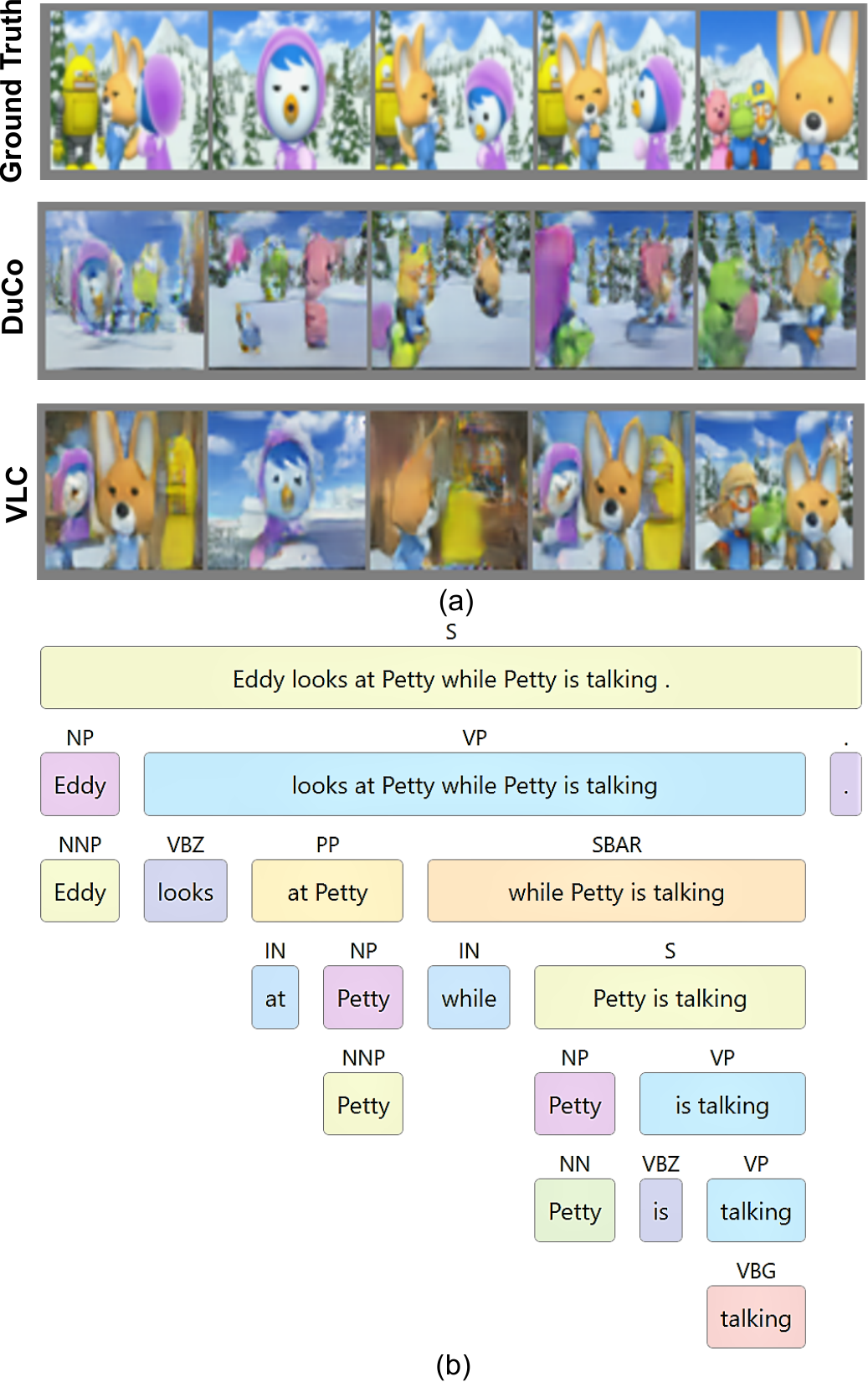}
\vspace{-10pt}
\caption{(a) Comparison of predictions from DuCo-StoryGAN and \modelname{}. (b) The constituency parse for caption of the fourth frame in (a). \vspace{-3pt}
\label{fig:const_example}}
\vspace{-15pt}
\end{figure}

\begin{figure}[t]
\centering
\includegraphics[width=0.48\textwidth]{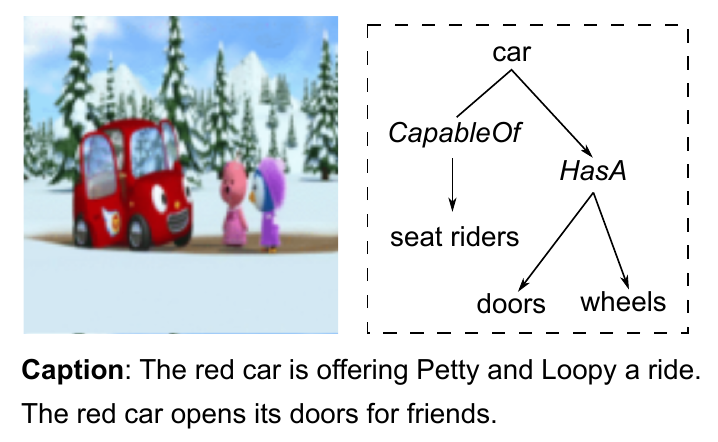}
\vspace{-5pt}
\caption{Example of commonsense knowledge for a given caption and its relevance to target image.
\label{fig:commonsense}}
\vspace{-5pt}
\end{figure}

\begin{figure}[t]
\centering
\includegraphics[width=0.47\textwidth]{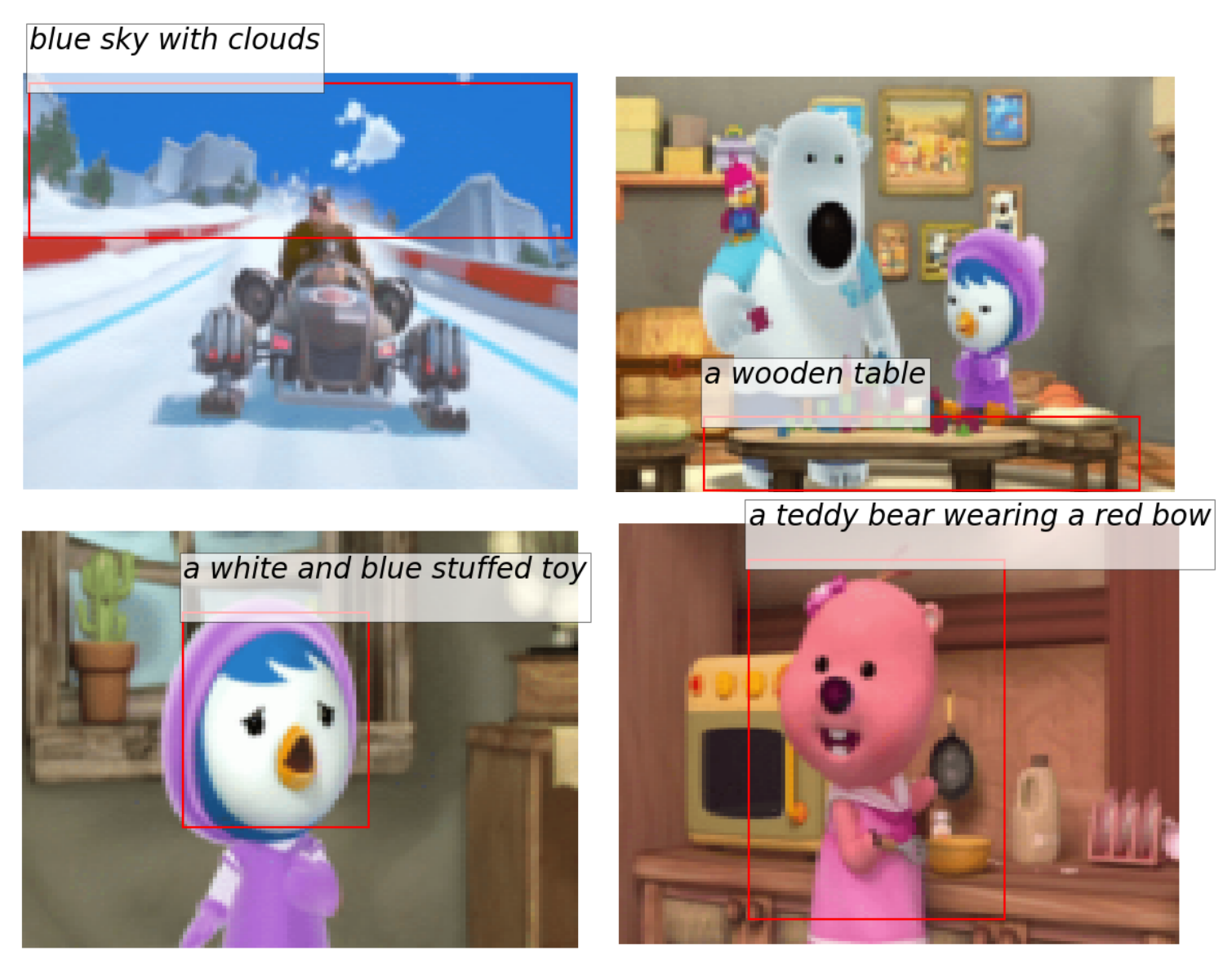}
\vspace{-5pt}
\caption{Dense captioning results on frames from the PororoSV  dataset.
\label{fig:densecap}}
\vspace{-7pt}
\end{figure}

\subsection{Linguistic \& Commonsense Knowledge}
Results from Table~\ref{tab:pororo_ablation} show that the grammatical structure of caption contributes to better understanding, which translates to improved visual stories. The improvement in frame accuracy further suggests that MARTT improves comprehension of multiple characters simultaneously present in the narrative. In order to further analyze this premise, we examine a story involving several characters and compare predictions from \modelname{} and DuCo-StoryGAN in Fig.~\ref{fig:const_example}. The constituency parse tree in Fig.~\ref{fig:const_example} shows the hierarchical understanding of the caption that is inherent in the MARTT architecture. Sub-tree masking allows the model to attend over multiple characters independently in earlier layers and combine the encoding in later layers. This semantic understanding is reflected in the image generated by \modelname{} which generates both characters mentioned in the caption distinctly, whereas Duco-StoryGAN barely generates one of them, validating the idea that grammatical knowledge is beneficial for story visualization.

In Fig.~\ref{fig:commonsense}, we demonstrate an example of commonsense knowledge for a single frame in a story. We extract a sub-graph containing general information about car from ConceptNet \cite{speer2017conceptnet} and use the graph contextualized embeddings from Graph Transformer for alignment with the generated image. The words \textit{door} and \textit{seat rider} correspond to specific sub-regions in the image and improve generalization.

\subsection{Analysis of Dense Caption Feedback}
We use the dense captioning predictions on ground truth images in the PororoSV dataset in order to obtain the dual learning loss signal for \modelname{} during training. While we expected the predictions to be noisy, we found many of the predictions to be surprisingly relevant to the PororoSV dataset. For instance, most of the characters in PororoSV were identified as \textit{teddy bear} or \textit{stuffed toy or animal} and the dense captioning model provides roughly accurate bounding boxes for the entire character or prominent body parts (see Fig.~\ref{fig:densecap}). This explains the improvement in character classification scores with the addition of dual learning via dense captioning in our model. Many of the background elements in the stories, such as \textit{blue sky}, \textit{wooden table}, \textit{snow}, and \textit{green tree} look similar to their realistic counterparts in our cartoon setting. The captions are usually missing descriptions as well as positions of these minute details, whereas the dense captioning model provides precise locations and descriptions for the same.

\subsection{Generalization to Flintstones Dataset}

\begin{figure}[t]
\centering
\includegraphics[width=0.47\textwidth]{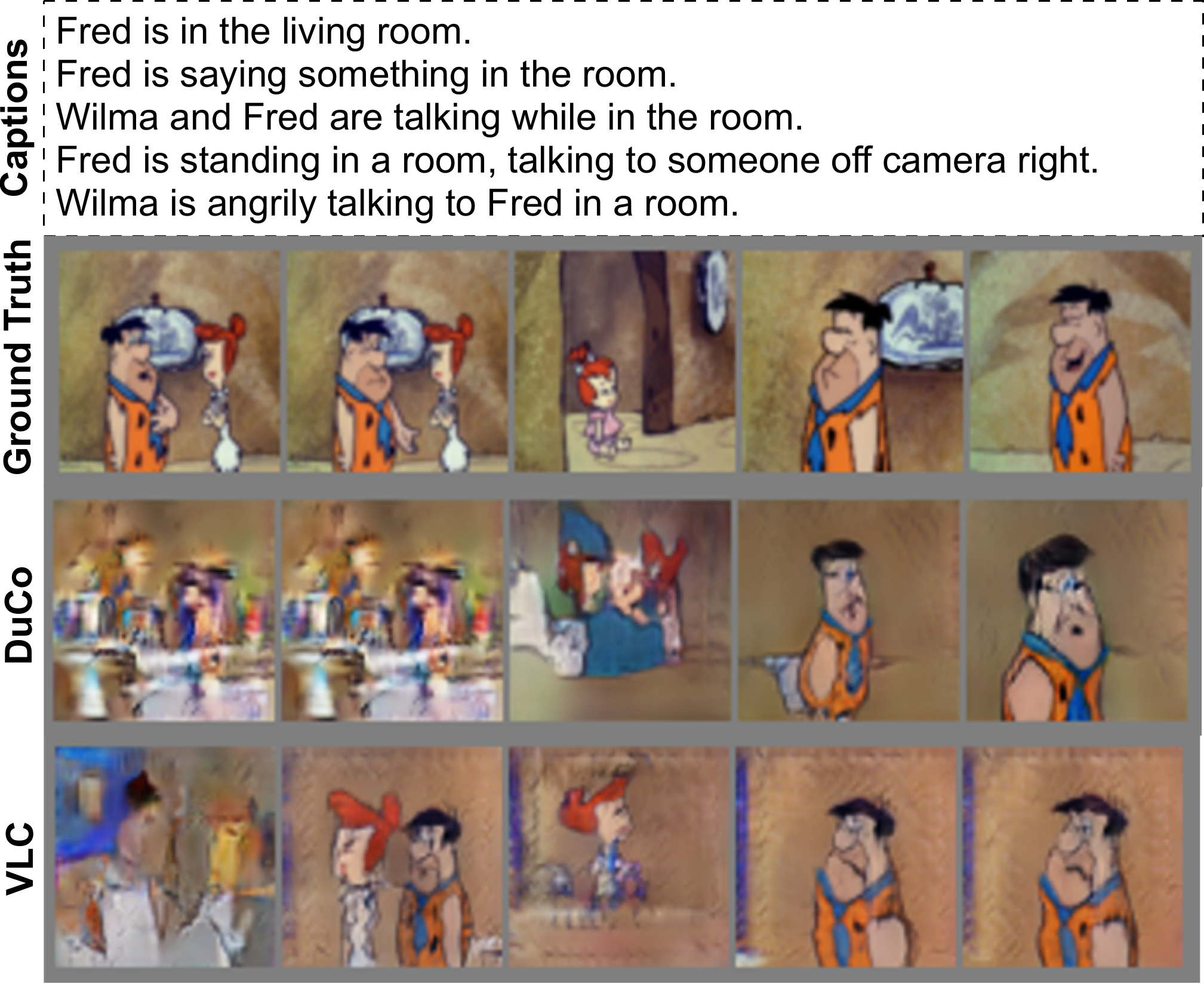}
\vspace{-5pt}
\caption{Initial results on the Flintstones dataset.
\label{fig:flintstones}}
\vspace{-7pt}
\end{figure}

In order to measure the generalization of our approach to another dataset, we transformed the Flintstones dataset presented in the text-to-video synthesis work, CRAFT \cite{gupta2018imagine}, into story visualization. A single frame is sampled from each video clip and frames from adjacent clips are gathered into stories of length 5 (similar to PororoSV). The resulting dataset, FlintstonesSV, has 7 major recurring characters and has 20132/2071/2309 samples in the training, validation and test splits. Our model \modelname{} outperforms DuCo-StoryGAN on all metrics. We see 3.89\% and 2.84\% improvements in character F1-score and frame accuracy with our structured framework. Additionally, the FID drops by 5.15\% suggesting large improvements in visual quality (see Fig.~\ref{fig:flintstones}). Under human evaluation, predictions from \modelname{} are preferred as much or more than those from Duco-StoryGAN for 90\% of the times. The \% of ties for all attributes is high, leaving scope for future research into this dataset.

\section{Conclusion}
In this paper, we investigate the use of structured knowledge for the task of story visualization. We propose a novel recurrent Tree-Transformer for encoding constituency trees and augment it with commonsense knowledge. We train the model using dense captioning loss and intra-story contrastive loss. Our results demonstrate the effectiveness of these approaches. We believe that these methods will encourage the use of structured knowledge for story visualization and text-to-image synthesis.

\section{Ethics/Broader Impacts}
\label{sec:ethics}
The datasets and corresponding train/validation/test splits used in this paper were proposed by \citet{li2019storygan}, \citet{pororvoQA}, \citet{gupta2018imagine} and \citet{maharana2021ducostorygan}. All the samples in the dataset consist of simple English sentences and cartoon images. Our experimental results are specific to the task of story visualization. The pretrained dense captioning model used in our paper is trained on English text and real-world images. All other models used and developed in our paper are trained on English text and cartoon images. By using cartoon images in our task, we avoid the egregious ethical issues associated with real-world usage of image generation such as DeepFakes. We focus not on generating realistic images, but on improved multi-modal understanding in the context of story visualization. 

\section*{Acknowledgments}
We thank Darryl Hannan, Hanna Tischer, Hyounghun Kim, Jaemin Cho, and the reviewers for their useful feedback. This work was supported by DARPA MCS Grant N66001-19-2-4031, DARPA KAIROS Grant FA8750-19-2-1004, ARO-YIP Award W911NF18-1-0336, and a Google Focused Research Award. The views are those of the authors and not of the funding agency.

\bibliography{emnlp2021}
\bibliographystyle{acl_natbib}

\appendix

\section{Methods}
\label{sec:methods}

\subsection{Dense Captioning}
For position invariance, we augment the PororoSV dataset with the mirror versions of the images and the corresponding mirror versions of the bounding box region predictions. When computing bounding box loss in dual learning, we compute the loss with the original bounding box prediction as well as its mirror version as target and retain the one which is lower. This way, we avoid penalizing the model for inverted positions of the characters since we do not provide explicit positional input to the model. 

\subsection{Story \& Image Discriminators} We use the story and image discriminators as outlined in StoryGAN \cite{li2019storygan}. The image discriminator is given the generated image $\hat{x}_k$, the sentence $s_k$, and the context information vector from the story encoder $h_0$, and distinguishes between a corresponding real triplet, containing the same information except for the real image $x_k$ instead of the fake image ($\mathcal{L}_{img}$). Additionally, the image discriminator also classifies the characters in the frame. The story discriminator evaluates the entire story $S$ and the generated sequence of images $\hat{X}$.

\subsection{Image Generation}
The image generator follows the two-stage approach in prior text-to-image generation works \cite{qiao2019mirrorgan,xu2018attngan,han2017stackgan, maharana2021ducostorygan}. The first stage uses outputs from the encoder; the resulting image is fed through a second stage, which weighs the outputs from the structure-aware context encoder as well as commonsense encoder, according to the image sub-regions and reuses for generation. The alignment module performs attention-based semantic alignment \cite{xu2018attngan} between image regions $h_k$ and words $\bar{m}_{k}=[f_{entity}(e_{k});f_{caption}(c_{k})]$ in the current timestep. $f_{entity}$ and $f_{caption}$ are dense layers for projecting commonsense and caption encodings respectively, into the same space as image embeddings. $\beta_{jik}$ indicates the weight assigned by the model to the $i^{th}$ word when generating the $j^{th}$ sub-region of the image. For the $j^{th}$ image sub-region, the word-context vector is calculated as:
\begin{gather*}
    a_{jk} = \sum_{i=0}^{L}\beta_{ji}\bar{m}_{ik};\>\> \beta_{jik}=\frac{\text{exp}(h_{jk}^{T}\bar{m}_{ik})}{\sum_{i=0}^{L}\text{exp}(h_{jk}^{T}\bar{m}_{ik})}
\end{gather*}

\section{Experimental Settings}
\subsection{Evaluation}
 \citet{li2019storygan} propose the character classification accuracy (exact match) within frames of generated visual stories as a measure of visual quality. \citet{maharana2021ducostorygan} propose an additional set of automated evaluation metrics that capture diverse aspects of a model's performance on visual story generation. We adopt those metrics for evaluating our models:
\begin{itemize}[leftmargin=*]
\itemsep0em 
    \item \textbf{Character Classification}: We use the finetuned Inception-v3 \cite{szegedy2016rethinking} and report frame accuracy and character F1-score.
    \item \textbf{Video Captioning Accuracy}: We use the pretrained MART video captioning model \cite{lei2020mart} and report BLEU2/3 scores for the generated captions.
    \item \textbf{R-Precision}: We use the Hierarchical-DAMSM \cite{maharana2021ducostorygan} to report R-Precision scores on the pairs of ground truth captions and generated stories.
    \item \textbf{Frechet Inception Distance (FID)}: We report the FID score, which is a metric used for evaluating the distance between real images and generated images for text-to-image synthesis datasets.

\end{itemize}

\subsection{Hyperparameters}
The image size that we use is 64-by-64, and the length of the story is 5 images/captions, same as DuCo-StoryGAN. The learning rates of the generator and discriminator are 2e-4. The model is trained for 150 epochs and the learning rate is decayed every 30 epochs. For each training update of the discriminators, two corresponding updates are performed for the generator network, with different mini-batch sizes for image and story discriminators \cite{li2019storygan}. The image discriminator batch size is 60 and the story discriminator batch size is 12. We found in our experiments that story visualization models are prone to mode collapse at lower batch sizes, which is not resolved with perceptual loss in contrast to conventional knowledge. The above-mentioned hyperparameters are optimized using 12 iterations of manual tuning.

The MARTT hyperparameters are as follows: The hidden size of the model is 192. The number of memory cells is 3. The number of hidden layers is 4. The dropout values across the model are 0.1. The layer normalization epsilon is 1e-12. The number of attention heads is 6. The word embedding size is 300 which is initialized using the 840B glove training checkpoint. The node embedding size is 50.

The total number of trainable parameters in the \modelname{} is approximately 100M. We use the ADAM optimizer with betas of 0.5 and 0.999.  We train the model on a single RTX A6000. Each epoch takes ~50 minutes, with the model being saved every 10 epochs. At 150 epochs of training, the total training time is nearly 4 days. 

\begin{figure*}[t]
\centering
\includegraphics[width=0.9\textwidth]{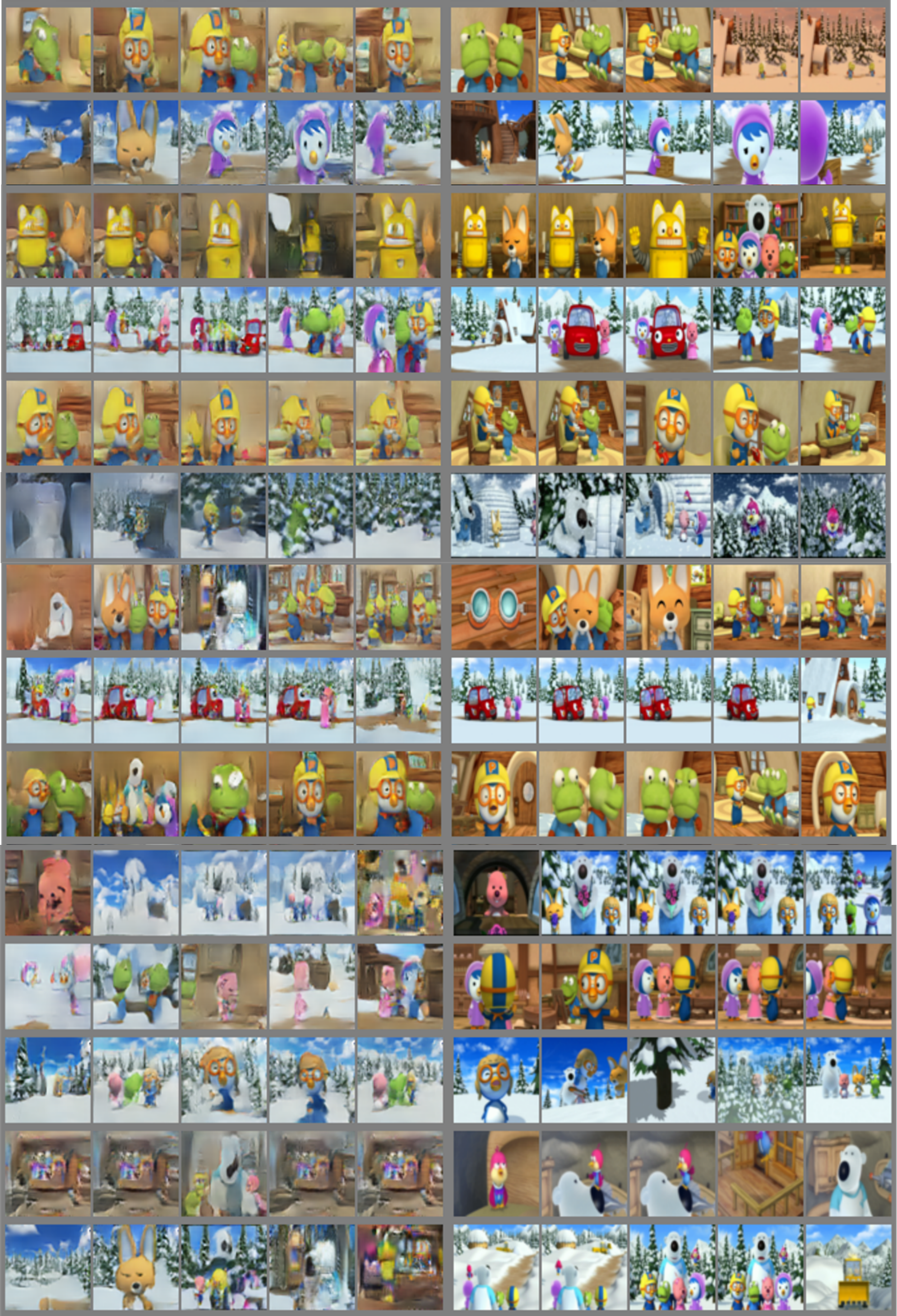}
\caption{Example of generated images (left) from \modelname{} and corresponding ground truths (right). 
\label{fig:example_1}}
\end{figure*}

\begin{figure*}[t]
\centering
\includegraphics[width=0.9\textwidth]{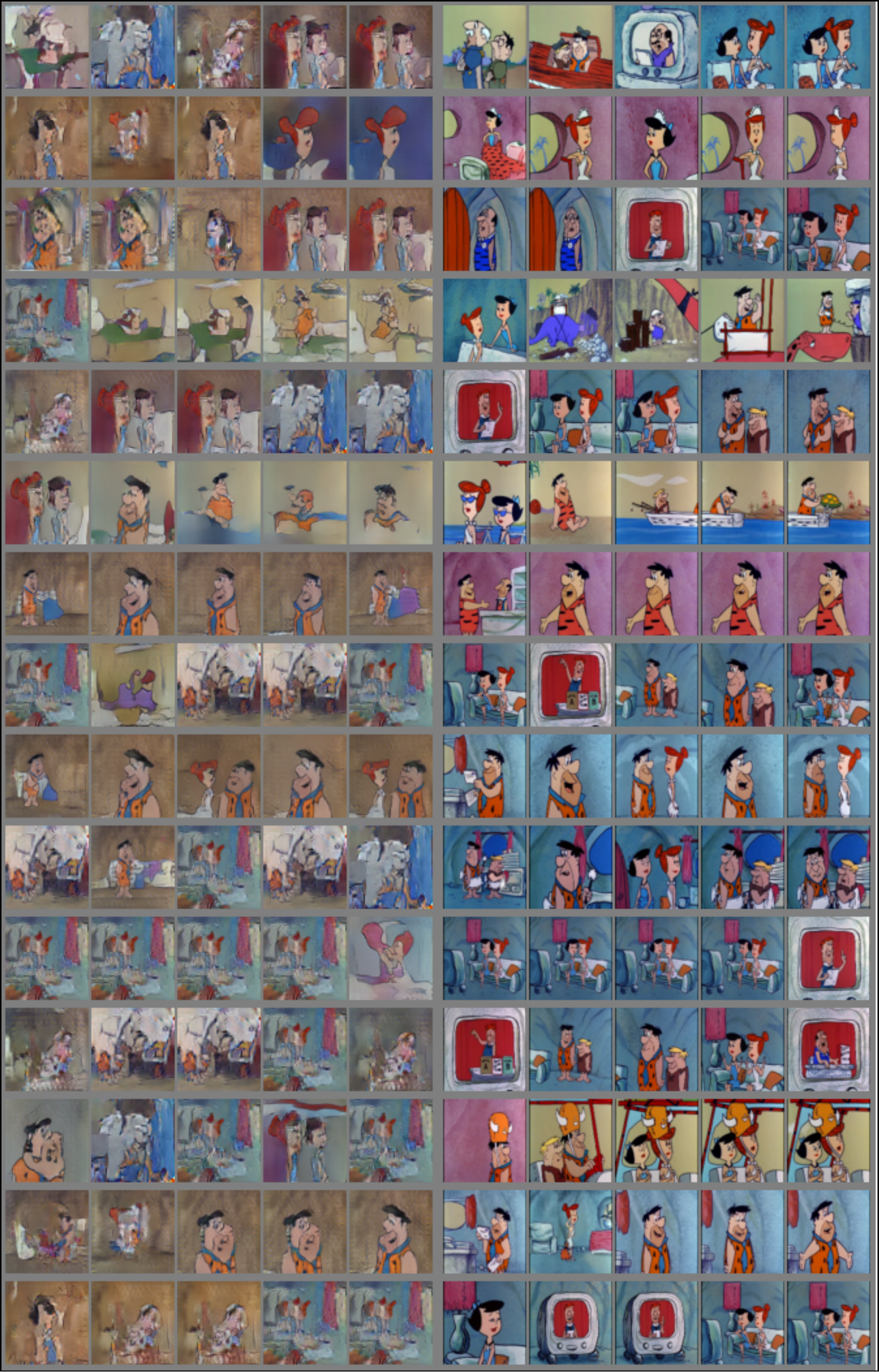}
\caption{Example of generated images (left) from \modelname{} and corresponding ground truths (right). 
\label{fig:example_2}}
\end{figure*}

\section{Results}

See examples of predictions for PororoSV and FlintstonesSV from \modelname{} in Figures~\ref{fig:example_1} and \ref{fig:example_2} respectively.

\subsection{Human Evaluation}
We conduct human evaluation on the generated images from \modelname{} and DuCo-StoryGAN, using the three evaluation criteria listed in \citet{maharana2021ducostorygan}: visual quality, consistence, and relevance. Two annotators are presented with a caption and the generated sequence of images from both models, and are asked to state their preferred sequence for each attribute. They also have the option to pick none if both images fare the same. In terms of visual quality, predictions from our model are preferred 62\% of the times, as compared to 28\% for DuCo-StoryGAN (see Win\% columns) for PororoSV. Our model is also preferred more times for the attributes consistency and relevance, but the higher \% of ties between the two models for these attributes indicate that much work remains to be done to improve global alignment between captions and images.

\end{document}